\documentclass[]{youtu} 

\usepackage{mathpazo}
\usepackage{graphicx}
\usepackage[numbers]{natbib}
\usepackage{times}
\usepackage[utf8]{inputenc} 
\usepackage[T1]{fontenc}    
\usepackage{url}            
\usepackage{booktabs}       
\usepackage{amsfonts}       
\usepackage{nicefrac}       
\usepackage{microtype}      
\usepackage{epsfig}
\usepackage{float}
\usepackage{multicol}
\usepackage{multirow}
\usepackage{amssymb}
\usepackage{amsmath}
\usepackage{wrapfig}
\usepackage{xspace}
\usepackage{color}
\usepackage{colortbl}
\usepackage[dvipsnames, svgnames, x11names, table]{xcolor}
\usepackage[misc]{ifsym}
\usepackage{makecell}
\usepackage{soul}
\usepackage{bm}
\usepackage{wrapfig}
\usepackage{subcaption, overpic, textpos}
\usepackage{paralist}
\usepackage{tabularx}

\usepackage{algorithm}
\usepackage{algorithmic}
\usepackage{adjustbox}
\usepackage{pifont}

\usepackage[capitalize]{cleveref}
\crefname{section}{Sec.}{Secs.}
\Crefname{section}{Section}{Sections}
\Crefname{table}{Table}{Tables}
\crefname{table}{Tab.}{Tabs.}

\newlength\savewidth

\renewcommand{\paragraph}[1]{\vspace{1.25mm}\noindent\textbf{#1}}

\definecolor{lightgray}{rgb}{0.8, 0.8, 0.8}
\definecolor{lgray}{rgb}{0.66, 0.66, 0.66}
\definecolor{whit_tab}{RGB}{255, 255, 255}
\definecolor{gray_tab}{RGB}{246, 246, 246}
\definecolor{oran_tab}{RGB}{252, 242, 237}
\definecolor{blue_tab}{RGB}{227, 240, 251}
\definecolor{lblu_tab}{RGB}{225, 235, 246}
\definecolor{orange_vitad}{RGB}{222, 131, 68}
\definecolor{blue_vitad}{RGB}{106, 153, 208}

\definecolor{trajectory_green}{RGB}{126, 171, 85}
\definecolor{trajectory_yellow}{RGB}{245, 194, 66}

\def\method{AdaKD}

\title{LLM-Oriented Token-Adaptive Knowledge Distillation} 

\author{
Xurong Xie$^1$\quad
Zhucun Xue$^1$\quad
Jiafu Wu$^2$\quad
Jian Li$^2$\quad
Yabiao Wang$^2$ \\
Xiaobin Hu$^2$\quad
Yong Liu$^1$\quad
Jiangning Zhang$^{1,2}$\quad
}
\affiliation{
$^1$Zhejiang University \quad
$^2$Tencent Youtu Lab \quad
}

\date{October 13, 2025}
\sourcecode{https://github.com/SassyRong/AdaKD}

\begin{document}

\abstract{Knowledge Distillation (KD) is a key technique for compressing Large-scale Language Models (LLMs), yet prevailing logit-based methods typically employ static strategies that are misaligned with the dynamic learning process of student models. 
These methods typically treat all tokens indiscriminately and apply a single, fixed temperature, resulting in suboptimal knowledge transfer. 
To address these limitations, we propose LLM-oriented token-\underline{\textbf{Ada}}ptive \underline{\textbf{K}}nowledge \underline{\textbf{D}}istillation (\textbf{AdaKD}), a novel framework that adapts the distillation process to the real-time learning state of each token. 
AdaKD consists of two synergistic modules driven by a unified token difficulty metric. 
First, our Loss-driven Adaptive Token Focusing (LATF) module dynamically adjusts the distillation focus by monitoring the student's learning stability, concentrating computational resources on the most valuable tokens at each training phase. 
Second, we introduce Inverse Difficulty Temperature Scaling (IDTS), a counterintuitive yet effective token-level temperature strategy. It employs low temperatures for difficult tokens for targeted error correction, and high temperatures for easy tokens to encourage the student to learn from the teacher's complete and smooth output distribution, thereby enhancing generalization. 
As a plug-and-play framework, AdaKD can consistently improve the performance of various distillation methods on multiple model architectures and benchmarks. 
}

\maketitle
\vspace{-.1em}

\section{Introduction} \label{sec:introduction}

Large Language Models (LLMs) have made significant advancements in recent years. They perform excellently on many natural language processing tasks, such as text generation, comprehension, and reasoning~\cite{achiam2023gpt,anil2023palm,grattafiori2024llama}. This success is mainly due to their \textit{extensive parameter sizes} and the pre-training they undergo on \textit{vast amounts of data}~\cite{kaplan2020scaling}. However, this powerful capability comes at the cost of \textit{enormous computational and storage resources}. These requirements create significant barriers to \textbf{\textit{deployment on edge devices in low-latency scenarios and to achieving widespread accessibility}}, limiting the practical reach of LLMs~\citep{wan2023efficient,zheng2025review,bai2024beyond}.

To solve above challenges, Knowledge Distillation (KD) has emerged as a promising solution for model compression and acceleration. Our work focuses on logit-based distillation, a prevalent white-box approach that directly transfers knowledge by matching the output distributions of the teacher and student models. While conceptually simple and effective, we argue that current logit-based methods still face two key limitations in adapting to the dynamic learning process of the student model: 

\textbf{\textit{1)} Indiscriminate token treatment.} Most methods treat all tokens indiscriminately, applying a uniform distillation objective across the entire sequence. This lack of differentiation is misaligned with the student's real-time learning progress, resulting in suboptimal knowledge transfer and potentially introducing noise from tokens that are already well-mastered.

\begin{figure}[t!]
    \centering
    \includegraphics[width=1\linewidth]{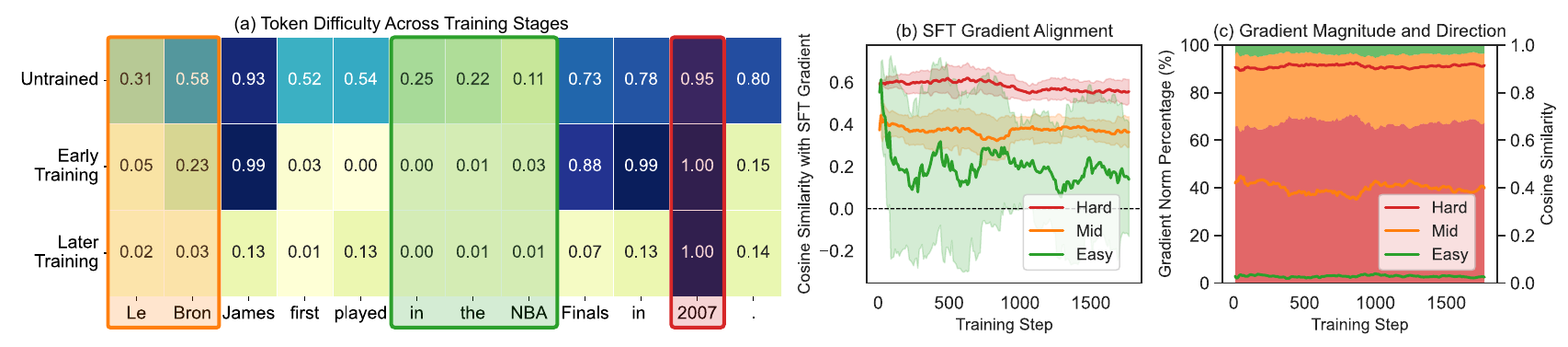}
    \caption{ Analysis of token difficulty and gradient dynamics. Tokens are grouped into \textit{Hard}, \textit{Mid}, and \textit{Easy} based on difficulty (Hellinger distance). 
    \textbf{(a)} Evolution of token difficulty across training stages. 
    \textbf{(b)} Cosine similarity of each token group's gradient with the SFT gradient. 
    \textbf{(c)} Each group's gradient norm percentage and its cosine similarity with the total batch gradient.}
    \label{fig:motivation}
\end{figure}

To better understand the consequences of this uniform treatment, we first investigate the learning dynamics of an instruction-following task at the token level (\cref{fig:motivation}). As shown in the \cref{fig:motivation}a, the difficulty of tokens for the student model is not static but evolves throughout the training process. Some tokens are persistently challenging (\textit{e.g.}, the token 2007, highlighted in the red box), requiring continuous focus. Others see their difficulty change dynamically (\textit{e.g.}, Le and Bron, in the orange box), while many "easy" tokens are quickly mastered in the early training stages (\textit{e.g.}, NBA and in, in the green box). This complex dynamic suggests that a static approach is suboptimal and motivates a token-wise, adaptive strategy.

Furthermore, we question the utility of continuing to train on "easy" tokens. We categorize tokens into "hard", "mid", and "easy" groups based on their difficulty and analyze their gradients. As shown in \cref{fig:motivation}c, easy tokens contribute negligibly to the parameter update, with their gradient magnitude being very small and their direction being nearly orthogonal to the overall batch gradient. More critically, \cref{fig:motivation}b reveals that the gradients of these easy tokens are unstable and poorly aligned with the supervised fine-tuning (SFT) direction, sometimes even moving in the opposite direction (negative cosine similarity). This evidence suggests that easy tokens provide limited learning value post-initial learning and may hinder knowledge transfer efficiency and stability via small, unstable gradients introducing conflicting signals.

To address above two limitations, we propose a novel Token-\textbf{Ada}ptive \textbf{K}nowledge \textbf{D}istillation (AdaKD) framework, which introduces a unified token difficulty metric driving two adaptive modules: 
\textbf{\textit{1)}} Loss-driven Adaptive Token Focusing (LATF) module dynamically selects the most valuable tokens for training at each stage, \textbf{\textit{2)}} while the Inverse Difficulty Temperature Scaling (IDTS) module taps temperature scaling's potential by assigning individual temperatures to tokens according to their learning difficulty.

In summary, our contributions are threefold.
\begin{itemize}
  \item We introduce a novel adaptive token selection mechanism that improves distillation efficiency by dynamically adjusting its focus based on the student model's learning stability.
  \item A novel token-level temperature scaling strategy that inversely correlates temperature with token difficulty to achieve both targeted error correction and enhanced generalization.
  \item Extensive empirical validation of AdaKD as a versatile and plug-and-play enhancement that consistently improves a variety of distillation baselines and architectures.
\end{itemize}
\section{Related Work} \label{sec:related_work}
\subsection{Knowledge Distillation for LLMs.}
Knowledge Distillation (KD) transfers knowledge from a large teacher to a smaller student. Methods are broadly divided into black-box and white-box distillation. 
Black-box approaches~\citep{yu2024distilling,hsieh2023distilling,ho2022large} use only the teacher's final outputs, making them suitable for closed-source models~\citep{achiam2023gpt, team2023gemini, IntroducingClaude} with less practical utility. 
Our work is in white-box distillation, which accesses teacher internals. Within this setting, feature-based distillation aligns intermediate hidden states, but this often requires complex, architecture-specific layer matching~\citep{sun2019patient, wang2020minilm, liang2023less}. Logit-based distillation, in contrast, offers a simpler approach by matching the final output distributions using a divergence measure. Beyond the foundational Forward KL Divergence (FKD)\citep{hinton2015distilling} and Reverse KL Divergence (RKD)\citep{minillm}, much recent work has focused on developing more advanced objective functions~\citep{wu2024rethinking, kodistillm, wang2025abkd}. Our approach is \textit{a plug-and-play framework that can be flexibly combined with these different objective functions}.

\subsection{Selective Token Distillation.}
Traditional KD treats all tokens equally, though not all tokens are equally informative~\citep{piantadosi2014zipf}. Consequently, many selective strategies have been proposed. One major direction is to focus the distillation loss on a subset of "important" tokens, identified based on metrics like difficulty or contribution to the teacher's prediction. For example, Selective Knowledge Distillation \cite{wang2021selective} uses a fixed ratio of tokens selected via a cross-entropy metric, which ignores the teacher's full distribution. A more advanced approach, AdaDS \cite{zhou2023adads}, dynamically adapts the difficulty metric (e.g., cross-entropy, confidence) using a lightweight RL selector. However, this strategy still relies on a pre-defined, fixed data selection ratio. Another line of work operates at the vocabulary level, for instance by preserving the relative order of top predictions~\citep{zhang2023towards, peng2025enhancing} or distilling only the top-k logits for efficiency~\citep{raman2023distillation, liu2024multi}. A common limitation in these approaches is the reliance on static or scheduled criteria. In contrast, \textit{our framework's LATF module dynamically adapts the token selection ratio itself based on the evolving training loss}, avoiding the static criteria of prior work.

\subsection{Adaptive Temperature Scaling.}
The distillation temperature is a key hyperparameter that modulates knowledge transfer by smoothing logits. Most methods employ a fixed temperature, which struggles to adapt to the student's evolving learning state. Consequently, dynamic temperature scaling has been well-explored, particularly in computer vision. Some strategies involve using different temperatures to normalize teacher and student logits~\citep{guo2022reducing, chi2023normkd}, while others adopt curriculum-based approaches that adjust the temperature to create an easy-to-difficult learning path~\citep{li2023curriculum}. However, such adaptive strategies are less common in LLM distillation. A representative work is Annealing KD~\citep{jafari2021annealing}, which lowers the temperature according to a predefined schedule. Such scheduled approaches are not adaptive to the model's real-time needs. Closer to our work in spirit, methods such as ATD~\cite{yang2025adaptive} and Mkdat~\cite{long2024mkdat} also adapt temperature according to the hardness of the sample. They typically rely on the cross-entropy loss to judge hardness and then use distinct functions to map this score to a temperature value. In contrast, \textit{our novel token-level IDTS module derives temperatures from a direct teacher-student discrepancy via a unique inverse scaling strategy}.

\section{Methodology} \label{sec:methodology}

\subsection{Preliminary of Knowledge Distillation in LLM}
Inference in Large Language Models (LLMs) is a sequential vocabulary classification task. Given a pair of prompt and target response, denoted as $(\mathbf{x},\mathbf{y})$, where $\mathbf{y}=(y_1,\ldots,y_L)$ is the target output sequence of length $L$, LLMs aim to predict the conditional probability distribution $p(\cdot|\mathbf{x},y_{<i})$ over the vocabulary $\mathcal{V}$ for each token $y_{i}\sim p(\cdot|\mathbf{x},y_{<i})$. KD minimizes the difference between the distributions predicted by the teacher $p$ and the student $q_{\theta}$ (parameterized by $\theta$). These distributions are obtained by applying a softmax function to the model output logits $z$, scaled by a distillation temperature $\tau$$:P(\cdot|\mathbf{x},y_{<i};\tau)=\mathrm{softmax}(z_P(\cdot|\mathbf{x},y_{<i})/\tau)$, where $P\in\{p,q_\theta\}$. The distillation loss is typically the average of token-level divergences computed using these temperature-scaled distributions, for each token $y_{i}$ in the ground-truth response $\mathbf{y}$. Thus, the classic FKD distillation loss~\citep{hinton2015distilling} is defined as:
\begin{equation}\label{eq:fkl}
\begin{aligned}
    &\mathcal{L}_{\mathrm{FKD}}=\frac{1}{L}\sum_{i=1}^{L}D_{KL}(p(\cdot|\mathbf{x},y_{<i};\tau)\parallel q_{\theta}(\cdot|\mathbf{x},y_{<i};\tau)).
\end{aligned}
\end{equation}
The KL divergence is computed over the vocabulary $\mathcal{V}$. Notably, the inclusion of temperature $\tau$ in the softmax function leads to a $\tau^2$  scaling factor in the final loss computation:
\begin{equation}\label{eq:kld}
\begin{aligned}
    D_{KL}(p\parallel q_{\theta})=\tau^{2}\sum_{y_i\in \mathcal{V}}p(y_i|\mathbf{x},y_{<i};\tau)\log\frac{p(y_i|\mathbf{x},y_{<i};\tau)}{q_{\theta}(y_i|\mathbf{x},y_{<i};\tau)}.
\end{aligned}
\end{equation}
Conversely, RKD loss~\citep{minillm} swaps the order of the distributions in the KL divergence, focusing on matching the modes of the teacher's distribution. These divergence measures form the basis of the distillation loss.

\subsection{{\method}: Token-\underline{Ada}ptive \underline{K}nowledge \underline{D}istillation} 
Building upon the insights from our analysis of token-level learning dynamics (\cref{fig:motivation}), we introduce Token-Adaptive Knowledge Distillation (AdaKD). Instead of a static approach, AdaKD is designed to dynamically tailor the distillation process—both its focus and intensity—to the real-time learning difficulty of each individual token. A detailed comparison of our framework with other relative methods is deferred to the appendix.

\begin{figure}[t!]
    \begin{minipage}[t]{0.5\textwidth}
        \vspace{0pt}
        \centering
        \includegraphics[width=\linewidth]{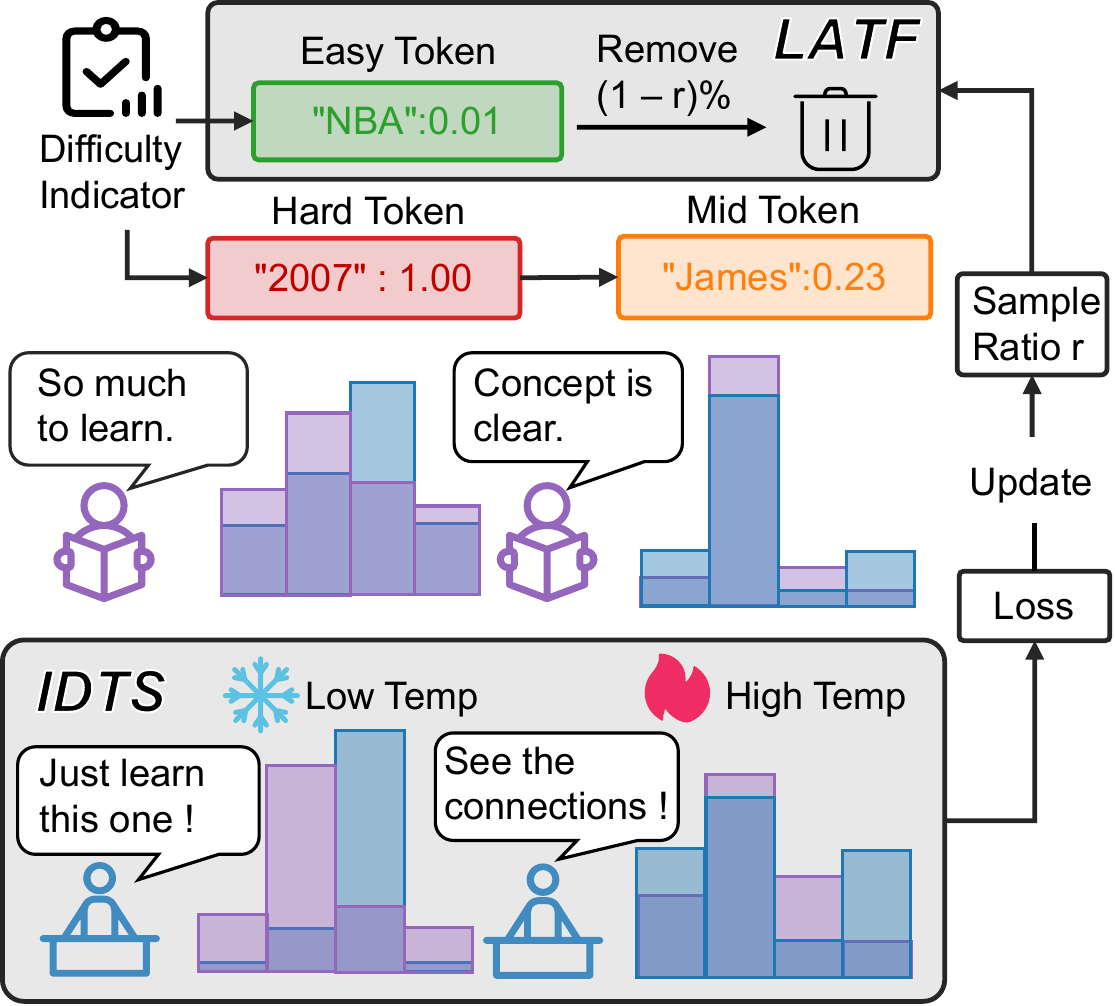}
        \caption{\textbf{Illustration of the AdaKD framework}. The bar charts visualize simplified teacher (blue) and student (purple) probability distributions. The top charts depict the initial learning gaps for "hard" and "mid-difficulty" tokens. After the LATF module filters tokens based on difficulty calculated via indicator, the IDTS module (bottom) applies low temperature to hard tokens for a sharp, corrective signal, and high temperature to easier tokens for a smoother distribution that enhances generalization.}
        \label{fig:framework}
    \end{minipage}\hfill
    \begin{minipage}[t]{0.46\textwidth}
        \vspace{0pt} 
        \begin{algorithm}[H] 
        \caption{Training Procedure of AdaKD.}
        \label{alg:AdaKD_concise}
        \begin{algorithmic}[1]
        \STATE \textbf{Input:} Teacher $p$, student $q_{\theta_0}$, dataset $\mathcal{D}$, total iterations $T$, temperature scale $c$, EMA decay rate $\beta$, tolerance $\epsilon$, step size $\delta$, warm-up steps $T_{\text{warmup}}$
        \STATE \textbf{Output:} Trained student model $q_{\theta_T}$.

        \STATE Initialize $t=1$, sample ratio $r_0=1.0$, compute initial loss $\bar{\mathcal{L}}_{0}$ with $q_{\theta_0}$, $\mathcal{L}_{\mathrm{ref}}=\infty$.

        \WHILE{$t < T$}
            \STATE Sample batch $(\mathbf{x}, \mathbf{y}) \sim \mathcal{D}$; Compute logits $\mathbf{z}_p, \mathbf{z}_{q_{\theta_t}}$
            
            \STATE Compute per-token difficulty scores $\mathbf{s}$ using Eq. \ref{eq:metric}
            
            \STATE Update focusing ratio $r_t$ using Eq. \ref{eq:LATF_update}
            
            \IF{$t > T_{\text{warmup}}$ and $r_t \neq r_{t-1}$}
                \STATE $\mathcal{L}_{\mathrm{ref}} \leftarrow \bar{\mathcal{L}}_{t-1}$. 
            \ENDIF
            
            \STATE Compute per-token temperatures $\tau$ using Eq. \ref{eq:token_wise_temp}

            \STATE $p\gets\mathrm{softmax}(\mathbf{z}_p/\tau)$

            \STATE $q_{\theta}\gets\mathrm{softmax}(\mathbf{z}_{q_{\theta_t}}/\tau)$.
            
            \STATE Compute $\mathcal{L}_{\text{AdaKD}}$ using Eq. \ref{eq:Selective}
            
            \STATE Update $\theta:\theta_{t} \gets \theta_{t-1}-\eta \cdot \nabla_{\theta_t} \mathcal{L}_{\text{AdaKD}}$

            \STATE $\bar{\mathcal{L}}_t = \beta \cdot \bar{\mathcal{L}}_{t-1} + (1-\beta) \cdot \mathcal{L}_{\text{AdaKD}}$
            
            \STATE $t \gets t+1$
        \ENDWHILE
        \end{algorithmic}
        \end{algorithm}
    \end{minipage}
\end{figure}

The entire AdaKD procedure is described in \cref{fig:framework}. Our framework is driven by two synergistic modules: Loss-driven Adaptive Token Focusing (LATF), which selects the most valuable tokens for training at each phase, and Inverse Difficulty Temperature Scaling (IDTS), which assigns a tailored temperature to each selected token. The foundation for both modules is a robust token difficulty indicator, which we will describe first.

\subsection{Choice of Difficulty Indicator} \label{sec:indicator}
The effectiveness of AdaKD depends on a metric that accurately quantifies token-level learning difficulty. We define this difficulty using the Hellinger distance~\citep{hellinger1909neue}, which measures the divergence between the teacher’s and student’s output probability distributions. For the i-th output token $y_i$, its difficulty score $s_i$ is calculated as:
\begin{equation}\label{eq:metric}
\begin{aligned}
s_i = \frac{1}{\sqrt{2}} \sqrt{\sum_{y_i \in \mathcal{V}} \left(\sqrt{p(y_i|\mathbf{x}, y_{<i})} - \sqrt{q_{\theta}(y_i|\mathbf{x}, y_{<i})}\right)^2}.
\end{aligned}
\end{equation}
The resulting score $s_i$ is bounded within the range of $[0, 1]$.
This indicator is chosen for its advantageous properties. First, its symmetry provides an unbiased measure of discrepancy, avoiding the inherent mode- or mean-seeking tendencies of asymmetric metrics like FKD and RKD. Second, its square-root operation compares the entire output distributions and is particularly sensitive to disagreements on low-probability candidates, thus providing a more comprehensive difficulty signal that captures subtle deviations in the student's replication of the teacher's full output distribution. 
\textit{This difficulty indicator $\mathbf{s}=(s_1,\ldots,s_L)$ then serves as the sole driving signal to synergistically guide the following two innovative modules that we designed.}

\subsection{Loss-driven Adaptive Token Focusing (LATF)} \label{sec:latf}
The gradient analysis in \cref{fig:motivation}b and \cref{fig:motivation}c reveals that training on "easy" tokens becomes inefficient and potentially unstable as training progresses. This strongly suggests that selectively focusing the distillation loss on a more valuable subset of tokens is beneficial. We implement this by applying the loss only to the top-$r\%$ of tokens with the highest difficulty scores:
\begin{equation}\label{eq:Selective}
\mathcal{L}_{\mathrm{distill}}=\frac{1}{L*r\%}\sum_{i=1}^LI_{r\%}(y_i)\cdot D_{KL}(q_{\theta}\parallel p), 
\end{equation}
where $L*r\%$ is the number of tokens that fall within the top-$r\%$ of the difficulty metric. The indicator function $I_{r\%}(y_i)$ is defined as:
\begin{equation}\label{eq:indicator}
I_{r\%}(y_i)=\begin{cases}1&\mathrm{if~}s_i\text{ ranks in the top }r\%\mathrm{~of~}\mathbf{s}\\0&\mathrm{otherwise}\end{cases} .
\end{equation}
\noindent However, using a fixed sample ratio $r$ is suboptimal. A static ratio cannot adapt to the model's changing learning state. To address this, we introduce LATF to adjust the focusing ratio $r_t$ dynamically. 
LATF operates via a simple feedback loop that monitors learning stability through the distillation loss. To obtain a stable signal, we first compute the exponential moving average (EMA) of the loss, denoted as $\bar{\mathcal{L}}_{t}$:
\begin{equation}\label{eq:ema}
\bar{\mathcal{L}}_t = \beta \cdot \bar{\mathcal{L}}_{t-1} + (1-\beta) \cdot \mathcal{L}_{\mathrm{distill},t}, 
\end{equation}
where $\beta$ is the decay rate of EMA and $\bar{\mathcal{L}}_{0}$ is the distillation loss when the student model is untrained. 
After an warm-up phase (where $r_t=1.0$), we set a loss reference point $\mathcal{L}_{\mathrm{ref}}$, which is initialized with the current EMA loss $\bar{\mathcal{L}}_t$. At each subsequent training step, LATF dynamically adjusts $r_t$ by comparing the latest $\bar{\mathcal{L}}_t$ to $\mathcal{L}_{\mathrm{ref}}$ within a tolerance $\epsilon$. Specifically, the update rules be described as:
\begin{equation}\label{eq:LATF_update}
r_t = \left\{
    \begin{aligned}
        & r_{t-1} \cdot (1 - \delta) &&\text{if } \bar{\mathcal{L}}_{t-1} < \mathcal{L}_{\mathrm{ref}} \cdot (1-\epsilon) \\
        & \min(1.0, r_{t-1} \cdot (1 + \delta)) &&\text{if } \bar{\mathcal{L}}_{t-1} > \mathcal{L}_{\mathrm{ref}} \cdot (1+\epsilon) \\
        & r_{t-1} &&\text{otherwise}, 
    \end{aligned}
\right.
\end{equation}
where $\delta$ is a small step size that controls the magnitude of adjustment. This rule creates an intuitive feedback loop.
We decrease the selection ratio $r_t$ to focus on more challenging tokens when the learning state is stable ($\bar{\mathcal{L}}_t$ drops below the lower bound). Conversely, we increase $r_t$ to incorporate simpler tokens for stabilization when the model struggles. The ratio remains unchanged within the tolerance zone to prevent over-reaction to normal training oscillations. After any adjustment to $r_t$, the reference point $\mathcal{L}_{\mathrm{ref}}$ is reset to the current $\bar{\mathcal{L}}_{t}$, keeping the performance baseline adaptive.

\subsection{Inverse Difficulty Temperature Scaling (IDTS)} \label{sec:idts}
Once LATF selects the tokens, IDTS determines the optimal temperature for distilling each one. Contrary to the conventional approach of using a high temperature to soften the teacher's distribution~\citep{jafari2021annealing}, we propose an inverse strategy: applying low temperatures to difficult tokens and high temperatures to easier ones.

Consider the information entropy \citep{shannon1948mathematical} of a probability distribution $\mathbf{p}=(p_1,\cdots,p_{\mathcal{V}})$, defined as $H(\mathbf{p})=-\sum_ip_i\ln(p_i)$, which quantifies the uncertainty of the distribution. The relationship between entropy and temperature can be precisely described by the derivative:
\begin{equation}\label{eq:derivative}
dH/d\tau=\operatorname{Var}_{p(\tau)}(z)/\tau^{3},
\end{equation}
where $\mathrm{Var}_{p(\tau)}(z)$ denotes the variance of logit $z$ under the distribution $p$ generated with temperature $\tau$. As variance is non-negative and $\tau > 0$, the derivative in Equation~\eqref{eq:derivative} is always non-negative, indicating that entropy is a monotonically increasing function of temperature.

Our IDTS module leverages this mathematical principle. For difficult tokens (high $s_i$), a low $\tau_i$ reduces the entropy, simplifying the learning objective into a sharp, corrective signal that focuses the student on matching the teacher's single best prediction. For easy tokens (low $s_i$), a high $\tau_i$ increases entropy, changing the objective to be more extractive . This encourages the student to learn the broader shape of the teacher's distribution, thereby enhancing generalization.

The implementation begins by converting the raw difficulty score $s_i$ into a normalized learning state $\hat{s}_i\in[-1, 1]$. This process is designed for robustness and stability: we first compute the ratio of $s_i$ to the batch median, chosen for its robustness to outliers. We then apply a $\log$ function to compress the long-tail distribution of these ratios, followed by a $\tanh$ function to smoothly map the result into the bounded range:
\begin{equation}\label{eq:relative_score}
\hat{s}_i = \tanh \left( \log\left(s_i/\text{median}(\mathbf{s})\right) \right).
\end{equation}
\noindent Subsequently, this learning state $\hat{s}_i$ dynamically modulates a base temperature $\tau_{\text{base}}$ via an exponential function:
\begin{equation}\label{eq:token_wise_temp}
\tau_i = \tau_{\text{base}} \cdot \exp(-c \cdot \hat{s}_i. \text{detach}()).
\end{equation}
\noindent Here, the negative sign enacts our inverse difficulty principle, with the hyperparameter $c$ controlling the modulation intensity. We chose this multiplicative approach because it makes the scaling effect robust to the specific value of $\tau_{\text{base}}$ and naturally constrains the final temperature $\tau_i$ to the predictable range of $[\tau_{base}\cdot e^{-c},\tau_{base}\cdot e^{c}]$. The entire calculation is detached from the computation graph, treating the resulting temperatures as fixed supervisory signals. The full AdaKD procedure, combining LATF and IDTS, is detailed in \cref{alg:AdaKD_concise}.

\subsection{Gradient Analysis of IDTS}
The loss function activates only high-difficulty tokens (where $I_{r\%}(y_i) = 1$). For these tokens, we compute the temperature $\tau_i$ scaling gradient of the KL divergence $D_{\text{KL}}(q_\theta \parallel p)$ with respect to student logits $z_q$:
\begin{align}
\frac{\partial D_{\text{KL}}^{(\tau_i)}}{\partial z_q(y_j)} &= \frac{1}{\tau_i} \left( q_\theta^{(\tau_i)}(y_j) - p^{(\tau_i)}(y_j) \right).
\end{align}
The update magnitude is governed by the gradient norm:
\begin{equation}
\left\| \nabla D_{\text{KL}}^{(\tau_i)} \right\|^2 = \sum_{y_j \in \mathcal{V}} \left( \frac{\partial D_{\text{KL}}^{(\tau_i)}}{\partial z_q(y_j)} \right)^2 = \frac{1}{\tau_i^2} \left\| q_\theta^{(\tau_i)} - p^{(\tau_i)} \right\|_2^2. 
\end{equation}
\noindent To minimize $\mathcal{L}_{\text{distill}}$, we need to maximize this gradient magnitude for accelerated convergence:
\begin{equation}
\min \mathcal{L}_{\text{distill}} \implies \max \sum_{\substack{i \\ I_{r\%}(y_i)=1}} \frac{1}{\tau_i^2} \left\| q_\theta^{(\tau_i)} - p^{(\tau_i)} \right\|_2^2.
\end{equation}
\noindent The difficulty metric $s_i$ is defined as the Hellinger distance:
\begin{equation}
s_i = \frac{1}{\sqrt{2}} \left\| \sqrt{p} - \sqrt{q_\theta} \right\|_2 \implies \left\| \sqrt{p} - \sqrt{q_\theta} \right\|_2^2 = 2s_i^2.
\end{equation}
\noindent Temperature scaling modifies the distribution discrepancy:
\begin{equation}
\left\| q_\theta^{(\tau_i)} - p^{(\tau_i)} \right\|_2^2 \propto \frac{1}{\tau_i^2} \left\| q_\theta - p \right\|_2^2. \quad
\end{equation}
\noindent Combining these relationships yields:
\begin{align}
\left\| q_\theta^{(\tau_i)} - p^{(\tau_i)} \right\|_2^2 \propto \frac{s_i^2}{\tau_i^2}, ~~~
\left\| \nabla D_{\text{KL}}^{(\tau_i)} \right\|^2 \propto \frac{1}{\tau_i^2} \cdot \frac{s_i^2}{\tau_i^2} = \frac{s_i^2}{\tau_i^4}.
\end{align}
\noindent Thus, the KL loss gradient is inversely related to the temperature \(\tau\). For difficult tokens, there is a larger discrepancy between the output distributions of the student and the teacher, student model require a larger gradient to approximate the teacher’s distribution, which corresponds to a lower temperature. For easy tokens, the output distributions of the student and teacher are more similar, the student model need a smaller gradient to prevent itself from diverging, which corresponds to a higher temperature.

\section{Experimental Results} \label{sec:experiment}

\subsection{Experimental Setups}
\paragraph{Datasets and Models.} Following the widely-adopted setup from \citet{minillm, kodistillm}, we use the databricks-dolly-15k dataset~\cite{conover2023free} for training and evaluate on five instruction-following benchmarks: Dolly-eval, Self-Instruct~\cite{wang2022self}, Vicuna-eval~\cite{chiang2023vicuna}, Super-Natural Instructions (S-NI)~\citep{wang2022super}, and Unnatural Instructions~\citep{honovich2022unnatural}. We demonstrate the generalizability of our framework on two modern model families: Qwen2-7B distilled to Qwen2-1.5B and OpenLLaMA2-7B to OpenLLaMA2-3B. 

\paragraph{Baselines and Implementation Details.} We compare AdaKD with supervised fine-tuning (SFT) and state-of-the-art KD methods, including FKD, RKD~\citep{minillm}, ABKD~\citep{wang2025abkd}, GKD~\citep{agarwal2024policy}, and DistiLLM~\citep{kodistillm}. For a fair comparison, all baselines are reproduced using their official implementations and meticulously tuned. 

Our experiments were conducted on a setup of 4 or 8 NVIDIA H20 80GB GPUs. For all distillation methods, we perform a hyperparameter search for the learning rate within the range of \{1e-4, 5e-4, 1e-5, 5e-5\} and for the global batch size within \{16, 32, 64, 128\}. Following the search, we set the learning rate to 5e-4 and the global batch size to 128 for all main experiments to ensure consistency. For the Qwen2 model, we use a batch size of 64 with a gradient accumulation factor of 2.

Following standard practice, we train the smaller GPT-2 model for 20 epochs, while the larger Qwen2 and OpenLLaMA2 models are trained for 10 epochs. For the Qwen2 and OpenLLaMA2 models, we employ Low-Rank Adaptation (LoRA) with a rank of 16 for parameter-efficient distillation. The final model checkpoints for evaluation are selected based on the highest ROUGE-L scores on the validation set.

\begin{table}[t!]
\centering
\footnotesize
\renewcommand{\arraystretch}{0.95}
\newcommand{\perf}[2]{#1{\tiny $\pm$#2}}
\newcommand{\perfbest}[2]{\textbf{#1}{\tiny $\pm$#2}}
\newcommand{\perfsecond}[2]{\underline{#1{\tiny $\pm$#2}}} %
\setlength{\tabcolsep}{6pt}
\caption{\textbf{Comparison of ROUGE-L scores for various KD methods on five instruction-following benchmarks.} All experiments were conducted using five different random seeds, with results reported as 'mean $\pm$ standard deviation'. For each student model configuration, optimal and sub-optimal results are highlighted in \textbf{bold} and \underline{underline}. 'w/ AdaKD' denotes our proposed plug-and-play enhancement, which \textbf{consistently improves performance across different base models}.}
\label{tab:main_results}
\begin{tabular}{@{}c l lccccccc@{}}
\toprule
\textbf{Model} & \textbf{Parameters} & \textbf{Method} & \textbf{Dolly} & \textbf{Self-Inst} & \textbf{Vicuna Eval} & \textbf{S-NI} & \textbf{UnNI} & \textbf{Avg.} \\
\midrule

\multirowcell{14}{\textbf{Qwen2} \\ ~\cite{yang2024qwen2}}
& \multirowcell{1.3}{\text{7B}} & Teacher & \perf{29.29}{0.56} & \perf{24.01}{0.63} & \perf{20.18}{0.72} & \perf{40.74}{0.63} & \perf{37.21}{0.76} & 30.29 \\
\cmidrule(l){2-9}
& \multirowcell{11}{\text{1.5B}} & SFT (LoRA) & \perf{24.82}{0.30} & \perf{18.79}{0.55} & \perf{17.99}{0.65} & \perf{32.13}{0.93} & \perf{31.04}{0.60} & 24.95 \\
\cmidrule(l){3-9}
& & FKD & \perf{25.72}{0.85} & \perf{20.13}{0.97} & \perf{18.24}{0.30} & \perf{35.77}{0.37} & \perf{32.93}{1.21} & 26.56 \\
& & \quad w/ AdaKD & \perf{25.94}{0.31} & \perf{19.75}{0.24} & \perf{18.36}{0.18} & \perf{35.88}{0.53} & \perf{33.21}{0.51} & 26.63 & ($\uparrow$0.07) \\
\cmidrule(l){3-9}
& & RKD & \perf{29.52}{0.50} & \perfbest{24.92}{0.66} & \perf{22.50}{0.51} & \perf{41.68}{0.67} & \perf{39.90}{0.49} & 31.70 \\
& & \quad w/ AdaKD & \perfsecond{30.03}{0.40} & \perfsecond{24.88}{0.71} & \perfsecond{22.97}{0.58} & \perfsecond{43.82}{0.78} & \perfbest{43.17}{0.32} & \textbf{32.97} & ($\uparrow$1.27) \\
\cmidrule(l){3-9}
& & ABKD & \perf{29.43}{0.60} & \perf{23.45}{0.63} & \perf{22.72}{0.60} & \perf{41.60}{0.79} & \perf{40.34}{0.47} & 31.51 \\
& & \quad w/ AdaKD & \perfbest{30.44}{0.50} & \perf{23.60}{0.75} & \perfbest{23.40}{0.77} & \perfbest{44.23}{1.39} & \perfsecond{42.54}{0.78} & \underline{32.84} & ($\uparrow$1.33) \\
\cmidrule(l){3-9}
& & GKD & \perf{27.13}{0.47} & \perf{20.89}{0.90} & \perf{19.41}{0.39} & \perf{38.25}{0.84} & \perf{35.01}{0.59} & 28.14 \\
& & \quad w/ AdaKD & \perf{27.98}{0.58} & \perf{23.00}{0.78} & \perf{19.62}{0.17} & \perf{40.31}{0.97} & \perf{37.77}{0.71} & 29.74 & ($\uparrow$1.60) \\
\cmidrule(l){3-9}
& & Distillm & \perf{29.10}{0.51} & \perf{22.92}{0.64} & \perf{21.79}{0.44} & \perf{41.26}{0.46} & \perf{38.80}{0.73} & 30.77 \\
& & \quad w/ AdaKD & \perf{29.69}{0.40} & \perf{23.55}{0.96} & \perf{22.11}{0.55} & \perf{42.91}{0.80} & \perf{40.73}{0.82} & 31.80 & ($\uparrow$1.03) \\
\midrule\midrule

\multirowcell{14}{\textbf{OpenLLaMA2} \\ ~\cite{geng2023openllama} }
& \multirowcell{1.3}{\text{7B}} & Teacher & \perf{28.16}{0.60} & \perf{20.40}{0.92} & \perf{17.62}{0.48} & \perf{30.45}{0.82} & \perf{33.18}{0.47} & 25.96 \\
\cmidrule(l){2-9}
& \multirowcell{11}{\text{3B}} & SFT (LoRA) & \perf{26.54}{0.13} & \perf{17.45}{0.42} & \perf{16.87}{0.27} & \perf{31.64}{0.88} & \perf{30.64}{0.49} & 24.63 \\
\cmidrule(l){3-9}
& & FKD & \perf{26.56}{0.38} & \perf{18.11}{0.60} & \perf{16.78}{0.40} & \perf{31.94}{0.79} & \perf{30.97}{0.52} & 24.87 \\
& & \quad w/ AdaKD & \perf{26.96}{0.58} & \perf{18.75}{0.55} & \perf{16.64}{0.47} & \perf{32.78}{0.92} & \perf{31.64}{0.65} & 25.35 & ($\uparrow$0.48) \\
\cmidrule(l){3-9}
& & RKD & \perf{29.13}{0.34} & \perf{20.08}{0.66} & \perf{19.49}{0.28} & \perf{35.20}{0.60} & \perf{37.60}{0.62} & 28.30 \\
& & \quad w/ AdaKD & \perfsecond{29.81}{0.35} & \perf{20.00}{0.55} & \perf{19.49}{0.37} & \perf{36.80}{1.13} & \perfbest{40.26}{0.54} & \underline{29.27} & ($\uparrow$0.97) \\
\cmidrule(l){3-9}
& & ABKD & \perf{29.45}{0.77} & \perfsecond{20.96}{0.76} & \perfbest{19.78}{0.26} & \perf{35.98}{0.74} & \perf{38.60}{0.63} & 28.95 \\
& & \quad w/ AdaKD & \perfbest{30.19}{0.50} & \perf{20.65}{0.32} & \perfsecond{19.55}{0.28} & \perf{36.38}{0.30} & \perf{39.82}{0.56} & 29.32 & ($\uparrow$0.37) \\
\cmidrule(l){3-9}
& & GKD & \perf{29.23}{0.41} & \perf{19.96}{0.80} & \perf{18.10}{0.75} & \perf{34.68}{0.58} & \perf{35.05}{0.63} & 27.40 \\
& & \quad w/ AdaKD & \perf{29.48}{0.15} & \perf{20.96}{0.56} & \perf{19.07}{0.32} & \perfbest{37.60}{0.43} & \perf{39.31}{0.27} & 29.28 & ($\uparrow$1.88) \\
\cmidrule(l){3-9}
& & Distillm & \perf{29.50}{0.56} & \perf{20.67}{0.86} & \perf{19.09}{0.44} & \perf{35.58}{0.66} & \perf{37.39}{1.13} & 28.45 \\
& & \quad w/ AdaKD & \perf{29.52}{0.63} & \perfbest{22.13}{0.47} & \perf{19.50}{0.50} & \perfsecond{37.23}{0.72} & \perfsecond{40.14}{0.71} & \textbf{29.70} & ($\uparrow$1.25) \\
\midrule\midrule\

\multirowcell{14}{\textbf{GPT-2}\\~\cite{radford2019language}}
& \multirowcell{1.3}{\text{1.5B}} & Teacher & \perf{28.17}{0.30} & \perf{15.93}{0.57} & \perf{16.99}{0.37} & \perf{29.08}{0.22} & \perf{34.53}{0.22} & 24.94 \\
\cmidrule(l){2-9}
& \multirowcell{13}{\text{0.1B}} & SFT & \perf{24.08}{0.42} & \perf{10.39}{0.53} & \perf{15.39}{0.16} & \perf{19.21}{0.73} & \perf{23.72}{0.40} & 18.56 \\
\cmidrule(l){3-9}
& & FKD & \perf{23.90}{0.65} & \perf{10.30}{0.15} & \perf{14.93}{0.32} & \perf{19.07}{0.22} & \perf{23.81}{0.33} & 18.40 \\
& & \quad w/ AdaKD & \perf{24.12}{0.36} & \perf{10.21}{0.49} & \perf{14.93}{0.44} & \perf{19.37}{0.63} & \perf{24.39}{0.79} & 18.60 & ($\uparrow$0.20) \\
\cmidrule(l){3-9}
& & RKD & \perf{26.05}{0.26} & \perf{12.24}{0.11} & \perf{15.89}{0.38} & \perf{25.24}{0.69} & \perf{31.05}{0.49} & 22.11 \\
& & \quad w/ AdaKD & \perf{26.11}{0.25} & \perf{11.89}{0.32} & \perf{15.84}{0.37} & \perf{26.61}{0.27} & \perf{33.29}{0.49} & 22.75 & ($\uparrow$0.64) \\
\cmidrule(l){3-9}
& & ABKD & \perf{26.01}{0.50} & \perf{12.30}{0.81} & \perf{15.90}{0.60} & \perf{27.99}{0.92} & \perf{32.35}{0.74} & 22.91 \\
& & \quad w/ AdaKD & \perf{26.00}{0.20} & \perf{12.48}{0.43} & \perf{15.08}{0.41} & \perfsecond{29.50}{0.33} & \perfsecond{35.12}{0.27} & 23.64 & ($\uparrow$0.73) \\
\cmidrule(l){3-9}
& & GKD & \perf{25.81}{0.47} & \perf{13.12}{0.57} & \perf{16.39}{0.17} & \perf{24.86}{0.58} & \perf{29.63}{0.31} & 21.96 \\
& & \quad w/ AdaKD & \perfbest{27.20}{0.25} & \perfsecond{13.34}{0.36} & \perfbest{17.63}{0.30} & \perf{28.29}{0.54} & \perf{33.92}{0.72} & \underline{24.08} & ($\uparrow$2.12) \\
\cmidrule(l){3-9}
& & Distillm & \perfsecond{26.81}{0.14} & \perf{12.56}{0.68} & \perfsecond{16.58}{0.20} & \perf{26.28}{0.71} & \perf{31.63}{0.82} & 22.77 \\
& & \quad w/ AdaKD & \perf{26.62}{0.51} & \perfbest{13.51}{0.49} & \perf{16.21}{0.29} & \perfbest{29.54}{0.46} & \perfbest{35.43}{0.80} & \textbf{24.26} & ($\uparrow$1.49) \\
\bottomrule

\end{tabular}
\let\perf\undefined
\let\perfbest\undefined
\end{table}

\paragraph{Evaluation.} We report the \textbf{ROUGE-L}~\citep{lin2004rouge} score to measure the quality of generated text. Following standard practice, we generate responses from all models with the decoding temperature and top-p both set to 1.0. To ensure statistical robustness, we use five different random seeds: \{10, 20, 30, 40, 50\} and report the averaged ROUGE-L scores.

We also employ a powerful large language model, Qwen3-32B, as an impartial judge to perform pairwise comparisons between the responses generated by a baseline method and its enhanced version with AdaKD. To mitigate position bias, the order of the two responses is swapped in a second evaluation round, and only consistent judgments are retained. The results, presented as win/tie/loss percentages for AdaKD, are reported on the Dolly-eval, Self-Instruct, and UnNI benchmarks.

\begin{table}[t!]
\centering
\small
\caption{LLM-as-a-Judge evaluation results for the Qwen2-1.5B distillation task. The table presents the win, tie, and loss rates (\%) of models enhanced with AdaKD against their respective baselines.}
\label{tab:llm_judge_results}
\setlength{\tabcolsep}{6pt}
\begin{tabular}{l ccc ccc ccc}
\toprule
& \multicolumn{3}{c}{\textbf{Dolly}} & \multicolumn{3}{c}{\textbf{S-NI}} & \multicolumn{3}{c}{\textbf{UnNI}} \\
\cmidrule(lr){2-4} \cmidrule(lr){5-7} \cmidrule(lr){8-10}
\textbf{Method} & Win\% & Tie\% & Loss\% & Win\% & Tie\% & Loss\% & Win\% & Tie\% & Loss\% \\ 
\midrule
FKD    & 18.80 & 59.00 & 22.20 & 19.42 & 54.25 & 26.33 & 16.96 & 65.14 & 17.90 \\
RKD    & 19.80 & 65.80 & 14.40 & 23.55 & 59.68 & 16.77 & 15.83 & 72.09 & 12.07 \\
ABKD   & 20.80 & 62.60 & 16.60 & 21.14 & 60.86 & 18.00 & 10.19 & 80.59 & 9.22 \\
GKD    & 20.00 & 64.20 & 15.80 & 22.43 & 62.75 & 14.82 & 18.30 & 69.60 & 12.10 \\
DistiLLM & 24.40 & 55.00 & 20.60 & 25.15 & 56.14 & 18.71 & 18.26 & 68.47 & 13.27 \\
\bottomrule
\end{tabular}
\end{table}

\subsection{Quantitative Results}
Table \ref{tab:main_results} validates AdaKD as a universal plug-and-play enhancement. While advanced objectives (\textit{e.g.}, RKD, ABKD) already outperform foundational methods and can even surpass complex Student-Generated Outputs (SGOs) based approaches(\textit{e.g.}, GKD),  AdaKD consistently elevates all of them to new state-of-the-art performance. This universal improvement demonstrates that dynamically adapting the distillation process to the student's real-time learning state is a robust and crucial element for effective knowledge transfer, offering a fundamental enhancement regardless of the underlying distillation objective.

Table \ref{tab:llm_judge_results} presents the results of the LLM-as-a-Judge evaluation. The findings reveal that the effectiveness of AdaKD varies between methods. Techniques like GKD and DistiLLM, which introduce SGOs that bring new instability, benefit significantly, as our LATF component filters the resulting noisy gradients for a more stable performance gain. In contrast, the improvement for FKD is marginal. This is attributed to a conceptual conflict: FKD's mean-seeking objective clashes with our IDTS's targeted correction for hard tokens.

\begin{table}[t!]
\centering
\caption{Comprehensive ablation studies of AdaKD (ROUGE-L scores). (a) Core components analysis. (b) Ablation on temperature scaling strategies.  (c) Comparison of different difficulty indicators. (d) Evaluation of LATF designs.}
\label{tab:ablation_all_vertical}

\newcommand{\best}[1]{\textbf{#1}}
\newcommand{\sbest}[1]{\underline{#1}}
\begin{minipage}[t]{0.45\textwidth}
    \vspace{0pt} 

    \begin{subtable}{\linewidth}
        \centering
        \caption{Core Components}
        \label{tab:ablation_core_3data}
        \setlength{\tabcolsep}{6.0pt}
        \renewcommand{\arraystretch}{0.95}
        \begin{tabular}{lcccc}
            \toprule
            \textbf{Method} & \textbf{Dolly} & \textbf{S-NI} & \textbf{UnNI} & \textbf{Avg.} \\
            \midrule
            RKD (Baseline) & 29.52 & 41.68 & 39.90 & 37.03 \\
            \quad + IDTS & \best{30.12} & \sbest{43.70} & \sbest{41.83} & \sbest{38.55} \\
            \quad + LATF & 29.50 & 41.88 & 39.82 & 37.07 \\
            \midrule
            AdaKD (Full) & \sbest{30.03} & \best{43.82} & \best{43.17} & \best{39.01} \\
            \bottomrule
        \end{tabular}
    \end{subtable}

    \vspace{1em} 

    \begin{subtable}{\linewidth}
        \centering
        \caption{Temperature Scaling}
        \label{tab:ablation_temp_short}
        \setlength{\tabcolsep}{6.0pt}
        \renewcommand{\arraystretch}{0.95}
        \begin{tabular}{l c c c c}
            \toprule
            \textbf{Method} & \textbf{Dolly} & \textbf{S-NI} & \textbf{UnNI} & \textbf{Avg.} \\
            \midrule
            $T=1.0$  & 29.50 & 41.88 & 39.82 & 37.07 \\
            \midrule
            AdaKD ($c=0.5$) & 30.03 & \best{43.82} & \best{43.17} & \best{39.01} \\
            Inv. Scaling ($-c$) & 28.93 & 40.02 & 38.06 & 35.67 \\
            \midrule
            $T=0.8$ & 30.00 & 41.41 & 40.10 & 37.17 \\
            $T=1.2$ & 29.55 & 42.07 & 40.05 & 37.22 \\
            $T \approx e^{-0.5}$ & \sbest{30.19} & \sbest{42.87} & \sbest{41.38} & \sbest{38.15} \\
            \midrule
            CTKD & \best{30.22} & 41.99 & 40.30 & 37.50 \\
            Logit Std. & 26.74 & 40.08 & 37.46 & 34.76 \\
            \bottomrule
        \end{tabular}
    \end{subtable}
\end{minipage}
\hfill 
\begin{minipage}[t]{0.51\textwidth}
    \vspace{0pt} 

    \begin{subtable}{\linewidth}
        \centering
        \caption{Difficulty Indicator}
        \label{tab:ablation_metric}
        \setlength{\tabcolsep}{4pt}
        \renewcommand{\arraystretch}{0.95}
        \begin{tabular}{l c c c c}
            \toprule
            \textbf{Method} & \textbf{Dolly} & \textbf{S-NI} & \textbf{UnNI} & \textbf{Avg.} \\
            \midrule
            RKD (Baseline) & 29.52 & 41.68 & 39.90 & 37.03 \\
            \midrule
            \multicolumn{5}{l}{\text{AdaKD with different metrics:}} \\
            + FKD & \sbest{30.29} & 42.81 & 42.34 & 38.48 \\
            + RKD & 29.93 & 43.02 & 42.23 & 38.39 \\
            + Cross-Entropy & \best{30.46} & 43.46 & 42.63 & \sbest{38.85} \\
            + JS-Divergence & 30.15 & 43.39 & 42.58 & 38.71 \\
            + NMTKD & 30.27 & \sbest{43.61} & \sbest{42.64} & 38.84 \\
            + Hellinger (Ours) & 30.03 & \best{43.82} & \best{43.17} & \best{39.01} \\
            \bottomrule
        \end{tabular}
    \end{subtable}

    \vspace{1em}

    \begin{subtable}{\linewidth}
        \centering
        \caption{LATF Design}
        \label{tab:ablation_latf}
        \setlength{\tabcolsep}{4pt}
        \renewcommand{\arraystretch}{0.95}
        \begin{tabular}{l c c c c}
            \toprule
            \textbf{Method} & \textbf{Dolly} & \textbf{S-NI} & \textbf{UnNI} & \textbf{Avg.} \\
            \midrule
            fixed $r=1.0$ & 30.12 & \sbest{43.70} & 41.83 & 38.55 \\
            \midrule
            LATF & 30.03 & \best{43.82} & \best{43.17} & \best{39.01} \\
            fixed $r=0.75$ & \sbest{30.27} & 43.49 & 42.02 & 38.59 \\
            linear $r: 1.0 \to 0.75$ & 29.63 & 43.37 & 41.83 & 38.28 \\
            cosine $r: 1.0 \to 0.75$ & \best{30.29} & 43.61 & \sbest{42.70} & \sbest{38.87} \\
            cosine $r: 1.0 \to 0.5$ & 29.74 & 42.71 & 41.74 & 38.06 \\
            \bottomrule
        \end{tabular}
    \end{subtable}
\end{minipage}

\end{table}

\subsection{Ablation Studies and Analyses}
We conduct ablation studies on the Qwen2-7B → Qwen2-1.5B distillation task using RKD as the baseline to dissect the contribution of each component in AdaKD. For clarity, the following tables detail results on Dolly, S-NI, and UnNI, the three benchmarks with the most extensive test items.

\paragraph{Impact of Core Components in AdaKD.}
\cref{tab:ablation_core_3data} reveals the synergy between our components. While integrating IDTS alone brings a substantial performance boost, LATF alone yields no improvement. This is consistent with our gradient analysis (Fig. \ref{fig:motivation}): LATF's primary role is to stabilize training by filtering out mastered tokens with unstable gradients, rather than directly advancing performance. The full AdaKD model achieves the best results, confirming a crucial synergy: LATF first removes noise to stabilize the learning process, which then allows IDTS to more effectively apply its adaptive teaching strategy to the remaining high-value tokens.

\paragraph{Analysis of the Difficulty Indicator.}
We evaluated several distribution metrics as the difficulty indicator, with results presented in \cref{tab:ablation_metric}. The results highlight that the optimal metric for an indicator differs from the distillation loss itself; for instance, FKL is a much more effective indicator than RKD; furthermore, symmetric metrics like Hellinger distance and JS-Divergence show a clear advantage over asymmetric ones on the large-scale S-NI and UnNI benchmarks. We also observe that Cross-Entropy, measured against the ground truth, performs best on the Dolly dataset, while NMTKD~\citep{zhang2023towards}, which focuses on aligning the top-k (k=5) predictions of each token, also demonstrates competitive performance. Ultimately, Hellinger distance achieves the highest average score, validating its use to provide the balanced and comprehensive disagreement signal crucial for our adaptive framework.

\begin{figure}[t!]
    \centering
    \includegraphics[width=\linewidth]{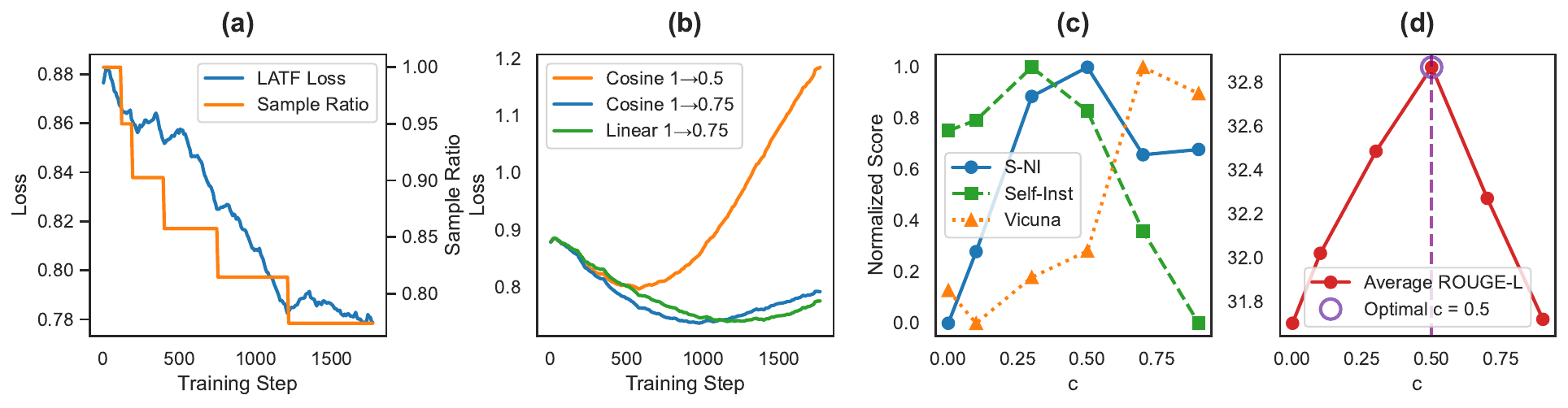}
    \caption{From left to right: (a) The loss and sample ratio of our adaptive LATF during training. (b) A comparison of loss curves for fixed scheduling strategies. (c) The effect of IDTS modulation intensity $c$ on normalized scores for individual datasets. (d) The average ROUGE-L score across datasets, showing the optimal choice for $c$.}
    \label{fig:latf_c}
\end{figure}

\paragraph{Analysis of LATF's Design.} We compare LATF against static and scheduled strategies in \cref{tab:ablation_latf}, using a target ratio $r=0.75$ for a fair comparison based on LATF's observed final value.  While these schedules prove competitive, LATF is ultimately more robust on challenging benchmarks. The reason is visualized in ~\cref{fig:latf_c}(a,b). For LATF (\cref{fig:latf_c}a), the sample ratio adapts to the training loss in real-time. The temporary increase in loss is an expected outcome of the model tackling a harder and more focused curriculum, a challenge it successfully overcomes. In contrast, scheduled methods (\cref{fig:latf_c}b) exhibit rising loss in later stages as they blindly enforce difficulty. This real-time adaptation makes LATF a more robust solution that eliminates the need for schedule-specific tuning.

The LATF module includes four hyperparameters. The most critical are the tolerance ($\epsilon$) and step size ($\delta$), which govern the feedback loop. As shown in Table \ref{tab:latf_sensitivity}, our grid search reveals stable performance when $\delta \leq \epsilon$, leading us to select $\epsilon=0.05$ and $\delta=0.05$ to balance responsiveness and stability. The other parameters, the EMA decay rate ($\beta$) and the warm-up ratio, were set to 0.97 and 0.05, respectively, to ensure a smooth training process.

\paragraph{Analysis of Key Design Choices within IDTS.}
\cref{tab:ablation_temp_short} validates IDTS's design against various temperature strategies. The failure of "Inverse Scaling" confirms our hypothesis that low temperatures are crucial for difficult tokens. However, simply using a low temperature globally is insufficient; AdaKD surpasses not only fixed-temperature baselines but also one using our method's optimal lower bound ($T \approx e^{-0.5}$). This proves the dynamic, token-level application is the key to success. Furthermore, AdaKD's superior performance over other adaptive methods like CTKD~\citep{li2023curriculum} and Logit Std~\citep{chi2023normkd, sun2024logit} highlights the effectiveness of our specific token-level design.

We analyze the impact of the IDTS modulation intensity $c$ in \cref{fig:latf_c}(c,d). While the optimal $c$ varies for individual datasets (\cref{fig:latf_c}c), the average performance across all benchmarks (\cref{fig:latf_c}d) robustly peaks at $c=0.5$. We therefore adopt this value for our main experiments.

\paragraph{Efficiency Comparison}
The additional computations in AdaKD for token difficulty and temperature have a negligible impact on training efficiency. This is because these steps are lightweight and detached from the computation graph, adding no overhead to backpropagation. Table \ref{tab:efficiency} quantitatively validates this by comparing the training throughput on the Qwen2 model.

\begin{table}[!t]

\begin{minipage}[b]{0.48\textwidth}
    \centering
    \caption{LATF sensitivity analysis on UnNI benchmark.}
    \label{tab:latf_sensitivity}
    \begin{tabular}{l ccc}
    \toprule
    \textbf{Step (\(\delta\))} & \(\epsilon = 0.02\) & \(\epsilon = 0.05\) & \(\epsilon = 0.1\) \\
    \midrule
    \textbf{0.02} & 42.74 & \textbf{43.23} & 42.93 \\
    \textbf{0.05} & 42.44 & 43.17 & 43.06 \\
    \textbf{0.1} & 41.73 & 42.12 & 42.94 \\
    \bottomrule
    \end{tabular}
\end{minipage}
\hfill
\begin{minipage}[b]{0.48\textwidth}
    \centering
    \caption{Training throughput (samples/sec) on Qwen2.}
    \label{tab:efficiency}
    \begin{tabular}{@{}lccc@{}}
    \toprule
    \textbf{Method} & \textbf{Baseline} & \textbf{w/ AdaKD} & \textbf{Change} \\
    \midrule
    RKD             & 8.21             & 7.94                  & -3.29\%    \\   
    GKD             & 1.56             & 1.55                  & -0.87\%    \\
    DistiLLM        & 3.88             & 3.82                  & -1.55\%    \\
    \bottomrule
    \end{tabular}
\end{minipage}

\end{table}

\begin{figure}[t!]
    \centering
    \includegraphics[width=1\linewidth]{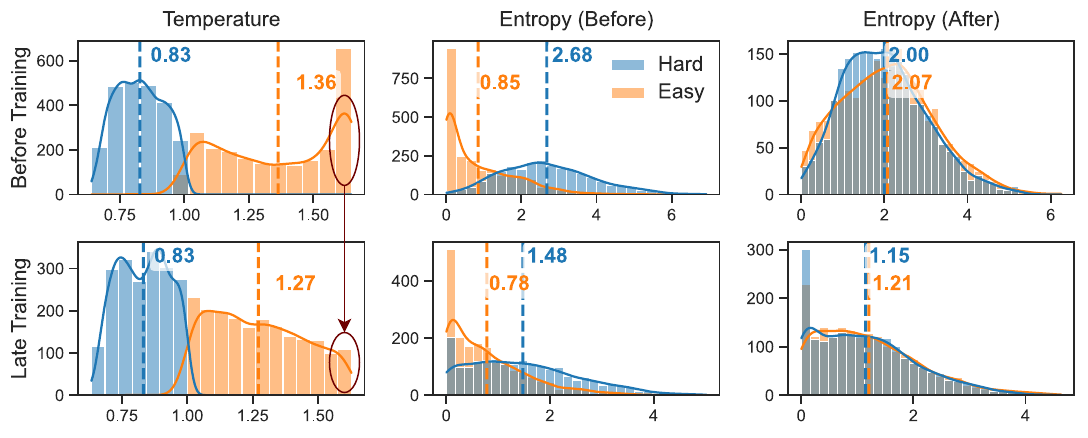}
    \caption{These histograms display the distribution of token counts (y-axis) across different metrics. The rows compare the model's state 'Before Training' (top) with 'Late in Training' (bottom). The columns, from left to right, show the distributions for assigned temperature, student’s output entropy before IDTS, and entropy after IDTS. Tokens are categorized into 'hard' (blue) and 'easy' (orange) groups, with dashed vertical lines indicating their respective means.}
    \label{fig:analyse}
\end{figure}

\noindent\textbf{Analysis of the Dynamic Mechanisms in AdaKD.} 
\cref{fig:analyse} illustrates the dynamic synergy of AdaKD's mechanisms by comparing distributions at the start and end of training.  A key observation is that our IDTS module consistently aligns the information entropy of each part of tokens, regardless of the training stage. This dynamically adjusts the learning objective for each token, guiding the model's output distributions toward a uniform level of uncertainty, regardless of their initial difficulty.

The temperature distribution reflects this adaptive strategy: IDTS consistently assigns lower temperatures to hard tokens and higher temperatures to easy ones. However, the distribution's evolution, highlighted in the red circles, reveals the critical synergy with LATF. Early in training, the temperature for easy tokens peaks sharply, as the "easy" set contains many trivial examples. Conversely, late in training, LATF has removed these mastered tokens, causing IDTS to assign a smoother range of high temperatures to the remaining, non-trivial "easy" set. This demonstrates how LATF dynamically refines the learning process, enabling IDTS to apply its scaling more effectively on tokens that still offer learning value.

\section{Conclusion} \label{sec:conclusion}
In this paper, we introduced Token-Adaptive Knowledge Distillation (AdaKD), a novel framework that dynamically adapts the distillation process to each token's learning state, overcoming the limitations of static distillation strategies. AdaKD synergistically combines Loss-driven Adaptive Token Focusing (LATF) to concentrate on valuable tokens and Inverse Difficulty Temperature Scaling (IDTS) to apply a highly effective temperature strategy for both error correction and generalization. Extensive experiments demonstrate that AdaKD, as a plug-and-play enhancement, consistently improves the performance of various distillation methods across multiple architectures. 

\noindent\textbf{Limitation.} Our work demonstrates the effectiveness of dynamically filtering tokens and dynamic temperature in distillation. Looking forward, there are clear paths for improvement.Currently, our Loss-driven Adaptive Token Focusing (LATF) module relies on a discrete adjustment mechanism, which can lead to slight training oscillations. Similarly, the formulation of our Inverse Difficulty Temperature Scaling (IDTS) module is based on heuristic principles and lacks a formal theoretical grounding. Future work can further explore the usage of continuous adjustment to obtain smoother convergence and more theoretical formulation like a direct control of the token-level information entropy for scaling. Lastly, our evaluation is limited to instruction-following tasks. Future work may expand the evaluation to challenging tasks including complex reasoning and code generation benchmarks.




\setcitestyle{numbers,square}

\bibliography{Tencent_Youtu_Lab/youtu_template}

\end{document}